\documentclass{article} 
\usepackage{iclr2015,times}
\usepackage{amsmath}
\usepackage{amssymb}
\usepackage{graphicx}
\graphicspath{{figures/}}
\usepackage{subfigure}
\usepackage{balance}
\usepackage{helvet}
\usepackage{courier}
\usepackage{color}
\usepackage{placeins}
\usepackage{hyperref}
\usepackage{url}
\usepackage{color}

\title{Predictive Encoding of Contextual Relationships for Perceptual Inference, Interpolation and Prediction}
\vspace{-0.5em}
\author{
Mingmin Zhao   \\
Computer Science Department\\
Peking University\\
\texttt{zhaomingmin@pku.edu.cn}
\vspace{-0.5em}
\And
Chengxu Zhuang \\
Electrical Engineering Department\\
Tsinghua University\\
\texttt{zcx11@mails.tsinghua.edu.cn}
\vspace{-0.5em}
\AND
Yizhou Wang   \\
Computer Science Department\\
Peking University\\
\texttt{Yizhou.Wang@pku.edu.cn}
\vspace{-0.5em}
\AND
Tai Sing Lee \\
Center for the Neural Basis of Cognition and Computer Science Department\\
Carnegie Mellon University\\
\texttt{tai@cnbc.cmu.edu}
}

\iclrfinalcopy 

\begin{document}
\vspace{-1em}
\maketitle

\vspace{-1em}
\begin{abstract}
\vspace{-0.2em}
\begin{quote}
We propose a new neurally-inspired model that can learn to encode the global relationship context of visual events across time and space and to use the contextual information to modulate the analysis by synthesis process in a predictive coding framework.
The model learns latent contextual representations by maximizing the predictability of visual events based on local and global contextual information through both top-down and bottom-up processes.
In contrast to  standard predictive coding models, the prediction error in this model is used to update the contextual representation but does not alter the feedforward input for the next layer, and is thus  more consistent with neurophysiological observations. We establish the computational feasibility of this model by demonstrating its ability in several aspects.
We show that our model can outperform state-of-art performances of gated Boltzmann machines (GBM) in estimation of contextual information.
Our model can also interpolate missing events or predict future events in image sequences while simultaneously estimating contextual information. We show it achieves state-of-art performances in terms of prediction accuracy in a variety of tasks and possesses the ability to interpolate missing frames, a function that is lacking in GBM.
\end{quote}
\vspace{-0.5em}
\end{abstract}
\vspace{-1em}

\section{Introduction}\label{sec:introduction}

In theoretical neuroscience, it has been proposed that in order to rapidly process  the constant influx of sensory inputs, which are complex, noisy and full of ambiguity, the brain needs to learn internal models of the world, and use them to generate expectations and predictions based on memory and context to speed up and facilitate inference.   Comprehension is achieved when the synthesized prediction or expectation, mediated by recurrent feedback from the higher visual areas to the early visual areas, explains the incoming signals \citep{mumford1992computational}. This framework  was recently popularized by \citet{rao1999predictive} in psychology and neuroscience as the predictive coding theory, and can be understood more generally in the framework of  hierarchical Bayesian inference  \citep{lee2003hierarchical}. The predictive coding idea has been generalized to non-visual systems \citep{bar2007proactive,todorovic2011prior}, and even a ``unified theory" of the brain \citep{friston2010free}. However, the computational utility and power of these conceptual models remains to be elucidated.


In this work, we propose a new framework that can learn internal models of  contextual relationships between visual events in space and time. These internal models of context allow the system to interpolate missing frames in image sequences or to predict future frames. The internal models, learned as latent variables, help accomplish these tasks by rescaling the weights of the basis functions represented by the neurons during the synthesis process  in the framework of predictive coding.  This model is inspired and related to Memisevic and Hinton's  \citep{roland_learn_to_relate,susskind2011modeling,memisevic2011gradient}  gated Boltzmann machines (GBM) and gated autoencoder (GAE) which also model spatiotemporal transformations in image sequences.  Their gated machines, modeling  3-way multiplicative interaction, make strong assumption on the role of  neural synchrony utilized in  learning and inference. Generalizing that model to $N$-way interaction is problematic because it involves $N+1$ way multiplicative interaction. As a result,  the GBM machines has  primarily been used to learn transformation between two frames. Prediction in GBM is accomplished by applying the transformation estimated based on the first two frames to the second frame to generate/predict a third frame.  It cannot interpolate a missing frame in the middle if given a frame before and  a frame after the missing frame.  Our model explicitly defines a  cost function based on mutual predictability. It is more flexible and can propagate information in both directions to predict future frames and interpolate missing frames in a unified framework.

In our formulation, synchrony is not required.  Evidence from multiple frames are weighed and then summed together in a way similar to spatiotemporal filtering of the input signals by visual cortical neurons in the primary visual cortex.  The inference of  the latent context variables is  nonlinear, accomplished by  minimizing the prediction error of the synthesized image sequence and the observed sequence. This is a crucial difference from GBM,  which, as is the case with most deep learning networks \citep{hinton2006fast},  relies on one-pass feedforward computation for inference.  Our model, by exploiting  top-down and bottom-up processes to minimize the same  predictive coding  cost function during both learning and inference, is able to estimate more meaningful and accurate contextual information.


Our framework of predictive coding under contextual modulation allows the model to accomplish the similar functions as GBM,  but also makes it more flexible and achieve more functions such as integrating more than 2 frames, and performing interpolation.  Our model is also more biologically plausible than the standard  predictive coding model \citep{rao1999predictive} in that the prediction error signals are used to update the contextual representation only, and do not replace the feedforward input to the next layer.  This model also provides a framework for  understanding how contextual modulation can influence the certain constructive and generative aspects of visual perception.

\section{Description of the Model}\label{sec:description_model}

The proposed model seeks to learn relationships between visual events in a spatial or temporal neighborhood to provide contextual modulation for image reconstruction, interpolation and prediction. It can be conceptualized as an autoencoder with contextual modulation, or a context-dependent predictive coding model. A predictive coding model states that the brain continually generates models of the world based on context and memory to predict sensory input.
It synthesizes a predicted image and feeds back to match the input image represented in the lower sensory areas.
The mismatch between the prediction and the input produces a residue signal that can be used to update the top-down models to generate predictions to explain away the inputs  \citep{mumford1992computational,rao1999predictive}. Our model extends this  ``standard'' predictive coding model, using the residue signals to update the contextual representation, which in turn  modulates the image synthesis process by rescaling  the basis functions of the neurons adaptively so that the synthesized images for all the frames in a temporal neighborhood maximally predict one another.

The problem can be formulated as the following energy function,
\begin{equation}\label{cost_with_sparsity}
L(\boldsymbol{x}_{1...N}, \boldsymbol{z};\boldsymbol{\theta}) = \sum_t{\|\boldsymbol{x}_t - \hat{\boldsymbol{x}}_t(\boldsymbol{x}_{1...N}, \boldsymbol{z};\boldsymbol{\theta})}\|^2_2 + \lambda\|\boldsymbol{z}\|_1,
\end{equation}
where $\boldsymbol{x}_{t}$ is the input signal, $\hat{\boldsymbol{x}}_{t}$ is the predicted signal, $\boldsymbol{z}$ is the contextual latent  variables,  $\boldsymbol{\theta}=\{\boldsymbol{W}_{1...N}, \boldsymbol{W}^z\}$ is a collection of parameters of the model to be learned, including the feedforward connections or receptive fields of the neurons in the $y$ hidden layer, and the feedback connections, $W^z$ from the  $z$ to modulate the generation of $\hat{x}$ by $y$. The second term in the function is a $L_{1}$ regularization term that makes the contextual latent variables sparse.  $\lambda>0$ serves to balance the importance of the prediction error term and the regularization term. Note that this cost function can be considered as a generalization of the classical energy functional or Bayesian formulation in computer vision (Horn 1986, Blake and Zisserman 1987, Geman and Geman 1984) with the first "data term" term replaced by a generative model and the second "smoothness prior" term replaced by a set of contextual relationship priors.

\begin{figure}[h]
    \centering
    \vspace{-2pt}
    \begin{minipage}[b]{0.45\textwidth}
    \centering
    \includegraphics[scale=1]{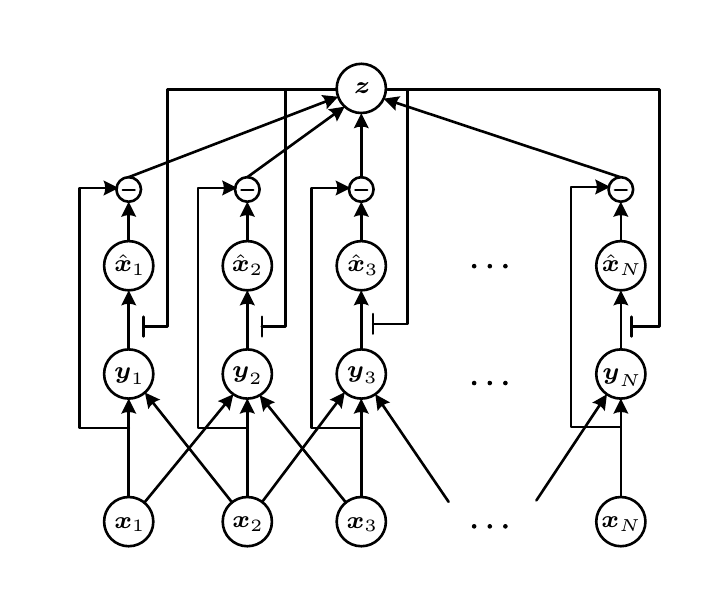}
    \end{minipage}
    \hfill
    \begin{minipage}[t]{0.54\linewidth}
    \centering
    \includegraphics[scale=1]{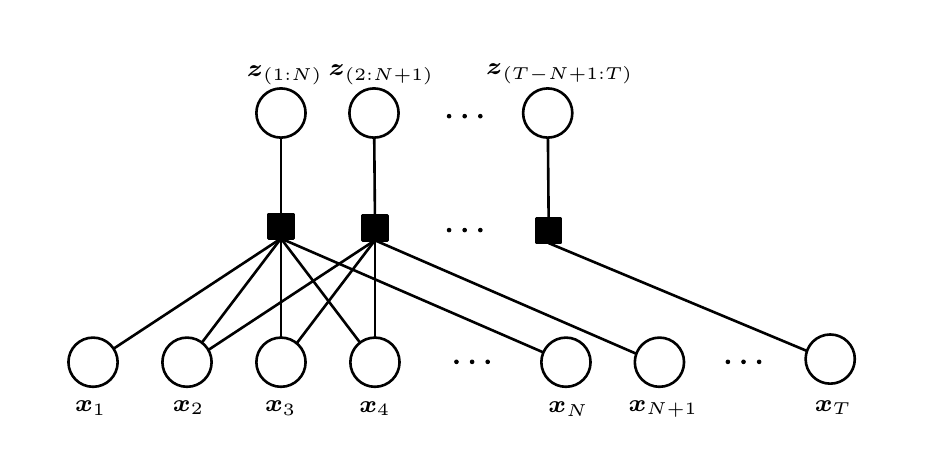}
    \end{minipage}
    \caption{Left: Computational circuit of the {\em Predictive Encoder}. Right: Graphical model representation of {\em Predictive Encoder} and its expansion for sequence of arbitrary length $T$.}\label{fig-predictive}
    \vspace{-0.3em}
\end{figure}

The objective of the model is to learn a set of relationship contexts that can modulate the predictive synthesis process to  maximize the mutual predictability of the synthesized images across space and/or  time. We will now describe how the prediction can be generated in this model. The model's information flow can be depicted as a circuit diagram shown in Figure \ref{fig-predictive} (left panel).  It consists of an input (visible) layer $\boldsymbol x_{t}$, a hidden layer $\boldsymbol y_{t}$ that performs a spatiotemporal filter operation on the input layer,
a prediction layer $\hat{\boldsymbol{x}}_{t}$ that represents the prediction generated by  $\boldsymbol y_{t}$  with  contextual modulation from the latent contextual representation layer $\boldsymbol z$. Prediction error signals are propagated up to drive the updating of the  contextual representation $\boldsymbol z$.

Units in the {\bf visible layer},
$\boldsymbol{x}_t\in\mathbb{R}^{D}, t = 1,...,N$,  represents a sequence of images, with  $\boldsymbol x_{t}$  representing the image frame (with $D$ number of pixels) at $t$ and $\boldsymbol{x}_{1...N}$ indicates a sequence of video image frames. Note that in the visible layer it is not necessary for all the visual events to be present, since the model possesses the ability to predict the missing events from available partial observations based on the principle of mutual predictability.

Units in the {\bf hidden layer},  $\boldsymbol{y}_t\in\mathbb{R}^{B}, t = 1,...,N$, are defined as:
\begin{equation}\label{eqn-hidden_y_t}
\boldsymbol{y}_t = \sum_{\tau\in\mathcal{N}(t)}{{\frac{1}{|\mathcal{N}(t)|}}\boldsymbol{W}_\tau\boldsymbol{x}_\tau},
\end{equation}
where $B$ is the number of $y_t$ units for each $t$, $\mathcal{N}(t)$ defines the index set of $\boldsymbol{x}_t$'s neighbors that provide the local temporal support to $y_t$, and $|\cdot|$ returns the size of a set. $\boldsymbol{W}_t \in\mathbb{R}^{B\times D}, 1\leq{t}\leq{N}$ is weight matrix to be learned as parameters. Each row of $\boldsymbol{W}_t$ can be viewed as a feature filter for a particular visual event in an image frame $t$. It can be considered as feedforward weight or filter for a hidden $y^{i}_{t}$ neuron or that neuron's spatial receptive field at a particular time frame. The corresponding rows of a particular sequence of related $W_t$ for unit $y^{i}_{t}$ is the {\bf spatiotemporal filter} receptive field of that neuron whose activity  $y^{i}_{t}$ as  the response of its spatiotemporal filter at frame $t$ to a particular sequence of image frames in $\mathcal{N}(t)$ .
Our definition of the neighborhood is flexible. The temporal neighborhood can be made causal, including up to frame $t$, or $t-1$ and the model would still work. Here, we use this non-causal symmetrical neighborhood to underscore the fact that our model can be used to model spatial context (which is symmetrical), and can go back and forth in time to modify our interpretation of the past events based on  current evidence and recent history. 

An additional crucial {\bf hidden layer},  with a set of latent variables $\boldsymbol{z}$, is used to model contextual  information. $\boldsymbol{z}$ is computed  by minimizing the residue errors between a sequence of  $N$ reconstructed image frames and the $N$ input frames. $\boldsymbol{z}$ is filtered by a weight matrix $\boldsymbol{W}^z$ (the dot product with  a row of this matrix) to provide feedback  to rescale the contribution of each latent variable activity $y^{i}_{t}$ for  generating the prediction signal ${\hat{\boldsymbol x}_{t}}$.     In Section \ref{subsec:evaluation_understand}, we will study the contextual representation $\boldsymbol{z}$ in greater details.

The prediction ${\hat{\boldsymbol x}_{t}}$ in the prediction layer is given by
\begin{equation}\label{eqn-recon_x_t}
\hat{\boldsymbol{x}}_t = {\boldsymbol{W}_t}^{\mathrm{T}}(\boldsymbol{y}_t\odot\boldsymbol{W}^z\boldsymbol{z}),
\end{equation}
 where $\boldsymbol{W}^z$ is a set of weights or basis functions that filter the contextual representation  $\boldsymbol z$ to generate a modulating signal for each $y_i(t)$,
$\odot$ is an element-wise product, and thus the contribution of each $y^{i}_{t}$ neuron to the predicted $\boldsymbol{\hat{x}}_{t}$ is its activity due to feedforward input rescaled by context modulation $\boldsymbol{W}^z\boldsymbol{z}$  to produce  a weight for its spatial  synthesis basis function ${\boldsymbol{W}_t}$.
Modulator $\boldsymbol{W}^z\boldsymbol{z}$ can be viewed as a high-dimensional distributed representation of context, the structure of which is modeled by a low-dimensional contextual representation $\boldsymbol{z}$ which is made sparse by the sparsity term.

Combining all the equations together,  the prediction generated by the context-dependent predictive coding model is given by
\begin{equation}\label{eqn-recon_x_given}
\hat{\boldsymbol{x}}_t(\boldsymbol{x}_{1...N}, \boldsymbol{z};\boldsymbol{\theta}) = {\boldsymbol{W}_t}^{\mathrm{T}}((\sum_{\tau\in\mathcal{N}(t)}{{\frac{1}{|\mathcal{N}(t)|}}\boldsymbol{W}_\tau\boldsymbol{x}_\tau})\odot \boldsymbol{W}^z\boldsymbol{z})
\end{equation}

Computationally, the update of the  contextual  latent variables $\boldsymbol z$ is driven by the residue signals $\boldsymbol{x}_{t} - \hat{\boldsymbol{x}}_{t}$.  The model can also be considered as a factor graph as shown in Figure \ref{fig-predictive} (right panel) together with its expansion for sequence of arbitrary length $T$. Each factor node (represented by the solid squares) corresponds to the mutual predictability defined by Equation \ref{cost_with_sparsity} between $N$ consecutive frames $\boldsymbol x_{t}$ modulated by the contextual representation  $\boldsymbol z_{(t:t+N-1)}$. Contextual representation  $\boldsymbol z$ will evolve over time, thus priors such as smoothness constraint can be imposed on the temporal evolution of $\boldsymbol z$, as shown in the graphical model, though no priors are imposed in our current implementation. At the abstract graphical model level, our model is very similar to autoencoder, as well as to Memisevic and Hinton's gated Boltzmann machines \citep{susskind2011modeling,memisevic2011gradient}. But there are difference in the concrete formulation of our model, as well as the learning and inference algorithms, with GBM, as discussed in the introduction.


\section{Description of the Algorithms}\label{sec:description_algorithms}
In this section, we describe the learning and inference algorithms developed for our model.   Bottom-up and top-down estimations are involved during both inference and learning.

\subsection{Unsupervised Parameters Learning}


The training dataset is composed of $m$ image sequences $\{\boldsymbol{x}^{(1)}_{1...N},...,\boldsymbol{x}^{(m)}_{1...N}\}$ and we assume each of them to be an i.i.d sample from an unknown distribution. The objective is to optimize the following problem:
\begin{equation}\label{optimization_learning}
\begin{aligned}
& \underset{\boldsymbol{\theta},\boldsymbol{z}}{\text{minimize}}
&& \sum_{i=1}^m{L(\boldsymbol{x}^{(i)}_{1...N}, \boldsymbol{z}^{(i)};\boldsymbol{\theta})},\\
\end{aligned}
\end{equation}

We adopt an EM-like algorithm that updates parameters and imputes hidden variable $\boldsymbol{z}$ alternatively while keeping the other one fixed.

\noindent{\bf Update $\boldsymbol{\theta}$}\hspace{2mm}
We use Stochastic Gradient Descent (SGD) to update $\boldsymbol{\theta}$ based on the following update rule:
\begin{equation}\label{eqn-update_rule1}
\boldsymbol{\theta}^{(k+1)} = \boldsymbol{\theta}^{(k)} + \Delta\boldsymbol{\theta}^{(k)}
\end{equation}
\begin{equation}\label{eqn-update_rule2}
\Delta\boldsymbol{\theta}^{(k)} = \eta\frac{\partial}{\partial\boldsymbol{\theta}^{(k)}}(\sum_{s\in S(k)}{L(\boldsymbol{x}^{(s)}_{1...N}, \boldsymbol{z}^{(s)};\boldsymbol{\theta}^{(k)})}) + \nu\Delta\boldsymbol{\theta}^{(k-1)}
\end{equation}
where the free parameter $\eta\in\mathbb{R}^{+}$ is the learning rate, $S(k)$ defines the mini-batch used for training at time $k$ and $\Delta\boldsymbol{\theta}^{(k-1)}$ is the momentum term weighted by a free parameter $\nu\in\mathbb{R}^{+}$. The momentum term helps to avoid oscillations during the iterative update procedure and to speed up the learning process. All the free parameters in the experiments are chosen under the guidance of \citet{hinton2010practical}.

The algorithm is implemented using Theano \citep{2010theano} which provides highly optimized symbolic differentiation for efficient and automatic gradient calculation with respect to the objective function. The idea of denoising \citep{vincent2008extracting} is also used to learn more robust filters. 

\noindent{\bf Estimate $\boldsymbol{z}$}\hspace{2mm}
Given fixed $\boldsymbol{\theta}$, we estimate the  contextual representation $\boldsymbol{z}^{(i)}$ for each sequence by solving the following optimization problem independently and in parallel:
\begin{equation}\label{optimization_update_h}
\begin{aligned}
& \underset{\boldsymbol{z}^{(i)}}{\text{minimize}}
\sum_t{\|\boldsymbol{x}^{(i)}_t - \hat{\boldsymbol{x}}_t(\boldsymbol{x}_{1...N}, \boldsymbol{z};\boldsymbol{\theta})\|_2^2} + \lambda\|\boldsymbol{z}^{(i)}\|_1\\
\end{aligned}
\end{equation}
where $\hat{\boldsymbol{x}}^{(i)}_t$ is computed by Eqn.\eqref{eqn-recon_x_given}.
To better exploit the quadratic structure of the objective function, we solve this convex optimization problem using a more efficient quasi-Newton method {\em Limited memory BFGS (L-BFGS)} algorithm instead of gradient descent \citep{ngiam2011optimization}.

During the training with each batch of data, we first update the parameter $\boldsymbol{\theta}$ using one step stochastic gradient descent, then iterate at most five steps of L-BFGS to estimate the hidden variable $\boldsymbol{z}$.
\subsection{Inference with Partial Observation}


Inference with partial observation refers to prediction or reconstruction of a missing image frame by the trained model given observed neighboring frames in the sequence. This problem is posed as an optimization problem that simultaneously estimates the latent variables $\boldsymbol{z}$ for the contextual representation and the missing event/frame $\boldsymbol{x}_u, 1\leq{u}\leq{N}$:
\begin{equation}\label{eqn-infer-optimization}
\begin{aligned}
& \underset{\boldsymbol{x}_u,\boldsymbol{z}}{\text{minimize}}
&& L(\boldsymbol{x}_{1...N}, \boldsymbol{z}; \boldsymbol{\theta})\\
\end{aligned}
\end{equation}
This optimization problem can be solved efficiently and  iteratively by an alternating top-down and  bottom-up estimation procedure. The top-down estimation ``hallucinates'' a missing event based on the neighboring events and the higher-level contextual representation. 
The bottom-up procedure uses the prediction residue error to update the contextual representation. Specifically, minimizing Eqn.\eqref{eqn-infer-optimization} is realized by alternately estimating $\boldsymbol{x}_u$ and $\boldsymbol{z}$ iteratively.

\noindent{\bf Estimate $\boldsymbol{z}$}\hspace{2mm}
Given learned $\boldsymbol{\theta}$ and current estimation of $\boldsymbol{x}_u$, we use the same method as Eqn.\eqref{optimization_update_h}. 

\noindent{\bf Estimate $\boldsymbol{x}_u$}\hspace{2mm}
Given learned $\boldsymbol{\theta}$ and current estimation of $\boldsymbol{z}$, we estimate a missing event/frame by solving the following optimization problem:
\begin{equation}\label{optimization_update_x_k}
\begin{aligned}
& \underset{\boldsymbol{x}_u}{\text{minimize}}
\sum_{t\in\mathcal{N}(u)\cup\{u\}}{\|\boldsymbol{x}_t - \hat{\boldsymbol{x}}_t(\boldsymbol{x}_{1...N}, \boldsymbol{z};\boldsymbol{\theta})\|_2^2}\\
\end{aligned}
\end{equation}
While Eqn.\eqref{eqn-recon_x_given} considers only the prediction of $\boldsymbol{x}_u$, this optimization problem factors the role of $\boldsymbol{x}_u$  in predicting/constructing its neighbors.
Notice that this objective function is a standard quadratic function, which has a closed form solution in one step.
For a video sequence, predicting a future frame and interpolating a missing frame are formulated and accomplished in a unified framework.

\section{Experimental Results}\label{sec:experiment}

\subsection{Receptive Field Learning}
In the first experiment, we trained our model using  movies synthesized from natural images. Each movie sequence exhibited either translation, rotation or scaling transformation.  We trained models for each type of transformation movies independently, as well as a mixture of the three.  We will show results of the feedforward filters $\boldsymbol{W}_t$ of models trained with three frames ($N=3$). The algorithm, however, is not limited to three frames and we will also show results of the model trained with relatively longer sequences such as six frames ($N=6$).

The images used to generate the training movie sequences were random samples from the whitened natural images used in \citet{olshausen1997sparse}. For translation or rotation, the patch size was $13 \times 13$ pixels. Translation steps were   sampled from the interval [-3, 3] (in pixels) uniformly, and  rotation angles were uniformly sampled from $[-21^\circ, 21^\circ]$ at  $3^\circ$ intervals. For scaling and the mixture of motion cases, the patch size was $21 \times 21$ pixels. The scaling ratio was uniformly sampled from $[0.6, 1.8]$. For mixture of motion, the training set was simply a combination of all three types of single-transformation movies, each with a constant transformation parameter. 
For models trained by a single type of motion, we used  25 contextual  representation units in $\boldsymbol z$ and 20000 training sequences and the size of $\boldsymbol{W}^z$ is $100 \times 25$.  For the models trained with all three types of motions, we used 75 contextual  representation units and 30000 training sequences and the size of $\boldsymbol{W}^z$ is $300 \times 75$.
We used unsupervised parameters learning algorithm described earlier with a learning rate ($\eta$) of 0.05 and momentum ($\nu$) 0.5. Every model was trained for 500 epochs.

\begin{figure*}[htb]
      \centering
      \vspace{-2pt}
        \subfigure[] {
        \includegraphics[scale=.551]{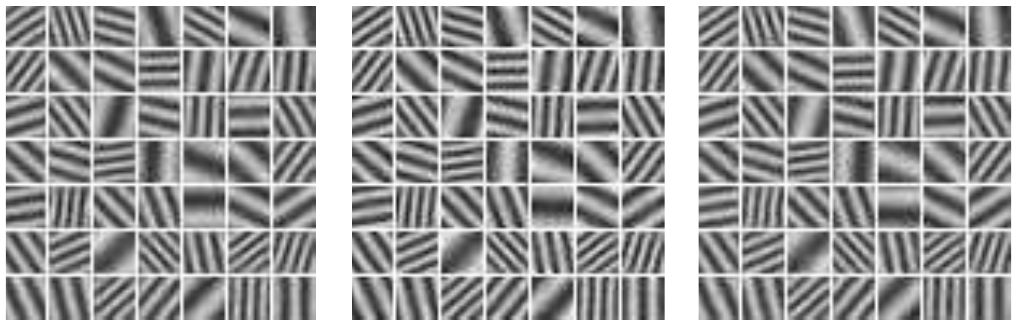}
        \label{filters_trans}
        }
        \subfigure[] {
        \includegraphics[scale=.551]{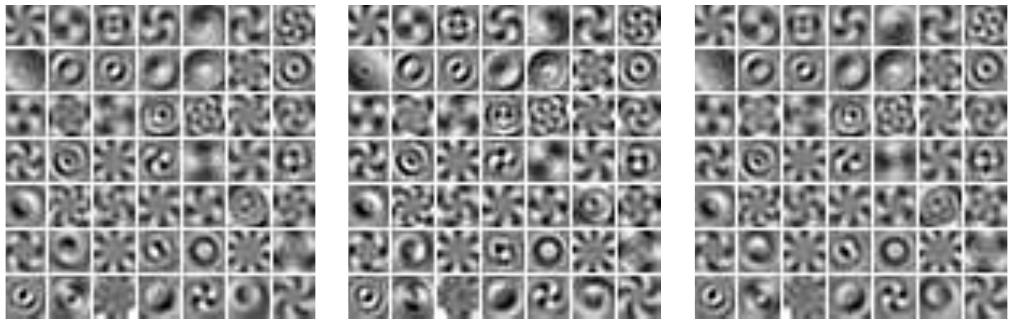}
        \label{filters_rotat}
        }
        \subfigure[] {
        \includegraphics[scale=.575]{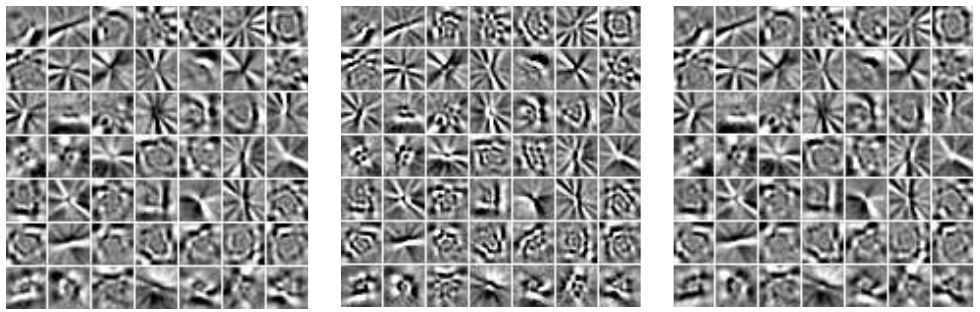}
        \label{filters_scaling}
        }
        \subfigure[] {
        \includegraphics[scale=.575]{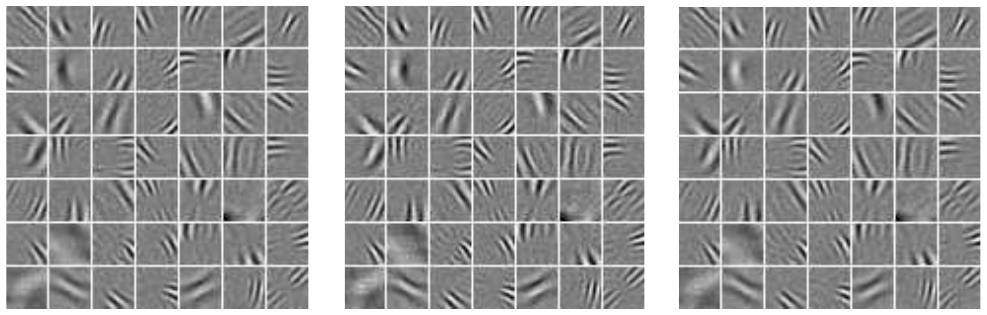}
        \label{filters_combine_sep}
        }
        \subfigure[] {
        \includegraphics[scale=1.65]{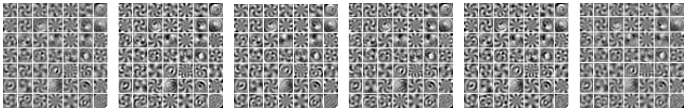}
        \label{filters_multi_rotat}
        }
        \vspace{-1em}
        \caption{Filters learned from: (a) Translation, (b) Rotation, (c) Scaling, (d) Mixture of Transformations, (e) 6-frame Rotation ($N=6$).}\label{fig_filter_learning}
        \vspace{-2pt}
\end{figure*}

Figure \ref{filters_trans} shows that the feedforward filters (or receptive fields) learned from translation resemble Fourier basis with a quadrature phase difference between frames.  Figure \ref{filters_rotat} shows that the filters learned from rotation are Fourier basis in polar coordinates, also with a quadrature phase in polar angle between frames.  The filters learned from scaling shown in Figure \ref{filters_scaling} depicts filters trained by scaling. They resemble  rays emanating from the center or circles contracting to the center,  reflecting  the trajectories of points during scaling. Figure \ref{filters_combine_sep} shows the filters trained with a motion mixture, which appear to encode the transformations in a distributed manner using localized Gabor filters, similar to the receptive fields of the simple neurons in the primary visual cortex.  
Figure \ref{filters_multi_rotat} shows the filters trained with $6$ frames rotation sequences.  This demonstrates the model can be used to learn longer sequence filters. We found training time scales linearly with $N$. 


\subsection{Understanding the Contextual Representation}\label{subsec:evaluation_understand}

To understand the information encoded in the contextual relationship latent variables $\boldsymbol z$, we used t-SNE method \citep{van2008visualizing} to see how pattern content and transformation content are clustered in a low-dimensional space. We applied the model pre-trained by the motion mixture to a combination of test data of 6,000 synthetic movies generated by randomly translating, rotating, or scaling image patches. The image patches were randomly sampled from 3 different datasets (MNIST, natural whiten images, and bouncing balls). Translation steps were no less than 1 pixel per frame, the rotation angles were no less than $6^\circ$, and scaling ratio sampled from $[0.6,0.9]$ or $[1.1, 1.8]$ to keep the different transformations distinct.

\begin{figure*}[htb]
      \centering
      \vspace{-1.5em}
        \subfigure[] {
        \includegraphics[scale=0.3]{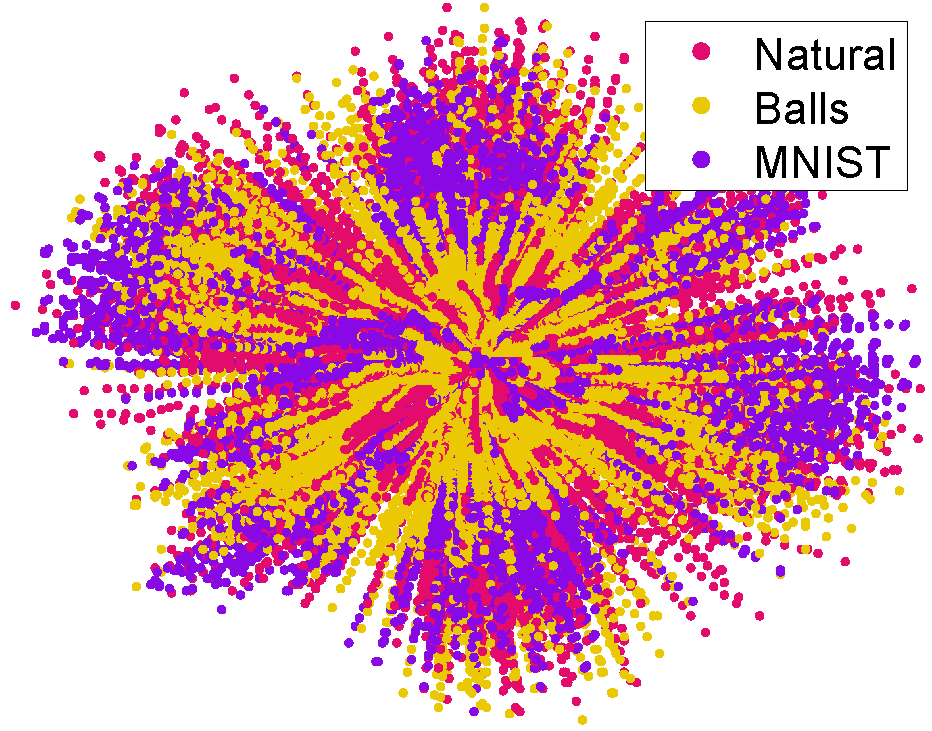}
        \label{fig_t-sne-visual}
        }
        \subfigure[] {
        \includegraphics[scale=0.3]{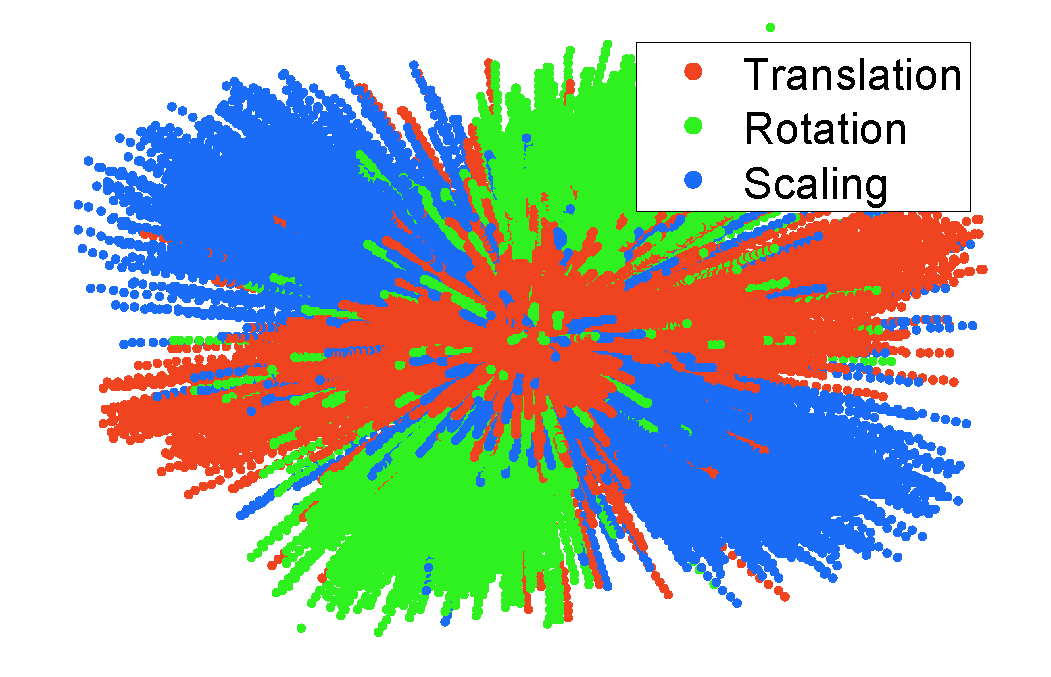}
        \label{fig_t-sne-visual_2}
        }
        \subfigure[] {
        \includegraphics[scale=0.2]{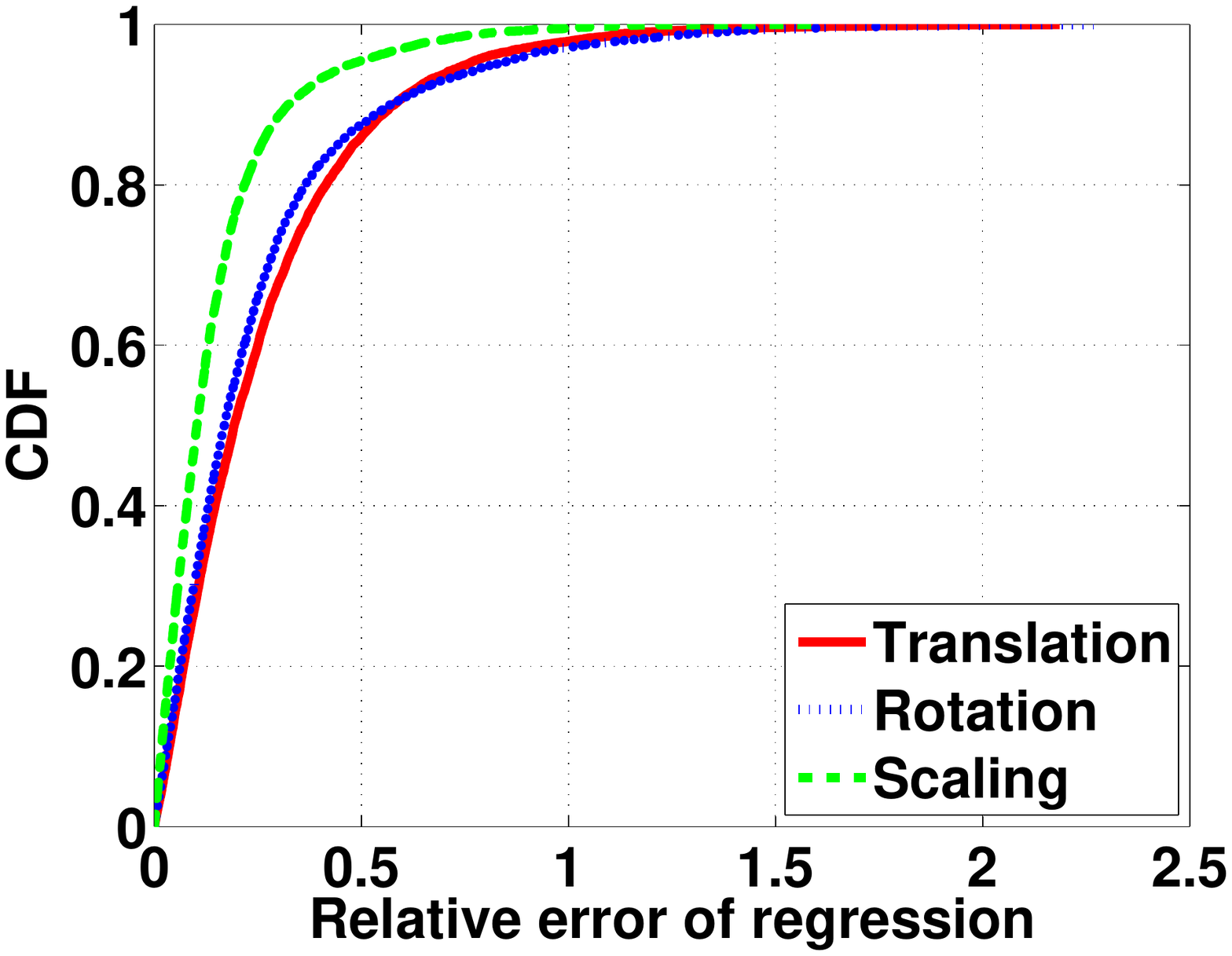}
        \label{fig_reg_error}
        }
        \vspace{-1em}
        \caption{a) t-SNE visualization for latent variables labeled by datasets; b) Visualization by transformations; c)CDFs of relative regression error}\label{fig_filter_learning}
        \vspace{-0.5em}
\end{figure*}

We visualize the activities in the z-layer using Hinton's  t-sne algorithm in Figure \ref{fig_t-sne-visual}, \ref{fig_t-sne-visual_2} in response to sequences from the three databases. It can be observed that the content data from the three databases (natural images, balls and MNIST) are all mixed together, indicating that the latent variables $\boldsymbol z$ cannot discriminate the image patterns. On the other hand, the transformations are relatively well clustered and segregated, suggesting that  these transformations are distinctly encoded in, and can be decoded from,  $\boldsymbol z$.


To investigate how transformations are distinctly represented in $\boldsymbol z$, we trained 3 SVM based on $\boldsymbol z$ to decode each of the three transformations (rotation, translation and scaling) from $\boldsymbol z$.   For each test sequence, we inferred the context representation $\boldsymbol{z}$, and then computed the probability of the three SVMs and chose the classification with the highest probability. All the SVMs were trained using the dot product as kernel function only.  The confusion matrix as shown in Table \ref{tab:svm_context_exist} suggests that the contextual representations encode the content-invariant transformation information.

\makeatletter\def\@captype{table}\makeatother
\begin{minipage}{.45\textwidth}
\centering
\begin{tabular}{|c|c|c|c|}
\hline
 & Our model & GBM & GAE\\
\hline
Accuracy & \textbf{0.9387} & 0.8664 & 0.7464\\
\hline
Time & 30 min & 2 hours & \textbf{15 min}\\
\hline
\end{tabular}
\caption{Prediction accuracy and training time (for 100 epoches).}
\label{tab:svm_context_accuracy}
\end{minipage}
\hfill
\makeatletter\def\@captype{table}\makeatother
\begin{minipage}{.45\textwidth}
\centering
\begin{tabular}{|c|c|c|c|}
\hline
 & Rotation & Translation & Scaling\\
\hline
Rotation & 9369 & 144 & 392\\
\hline
Translation & 427 & 9773 & 330\\
\hline
Scaling & 204 & 83 & 9278\\
\hline
\end{tabular}
\caption{Confusion Matrix: first column is predicted labels}
\label{tab:svm_context_exist}
\end{minipage}

We also compared the representational power of the inferred $\boldsymbol z$ in our model, computed using bottom-up and top-down processing, with the transformation latent variables $\boldsymbol z$ in the GBM and GAE, computed using one-pass feedforward computation. We compared a 2-frame version of our model with 3-way (2-frame) GBM and GAE \footnote{We used the CPU implementation of GBM on \url{http://www.cs.toronto.edu/~rfm/factored/} and used theano implementation of GAE on \url{http://www.iro.umontreal.ca/~memisevr/code.html} and ran it on CPU.} and trained 3 SVMs for each model to decode the type of transformation. As shown in Table \ref{tab:svm_context_accuracy}, our model is comparable, and in fact outperforms those two models while the time needed for training is comparable to GAE and faster than GBM.

In addition,  we trained a linear regression model using contextual representation $\boldsymbol z$ as the regressor to predict/estimate the transformations, namely, translation velocity, angular rotation velocity  and scaling factor. $\rho$ denotes the three transformation parameters. The relative regression error is defined as $\frac{|\rho_{groundtruth}-\rho_{predicted}|}{|\rho_{groundtruth}|}$. The cumulative distribution functions (CDFs) of relative regression error of the estimates in Figure \ref{fig_reg_error} shows that the contextual representation  contains sufficient information about the transformation parameters. 

\subsection{Prediction and Interpolation}

A crucial feature of our  model is its ability to predict and interpolate. Note that the GBM  can perform prediction but not interpolation. We first tested the model's ability in interpolation and prediction using training sequences generated from three datasets: face images \citep{gourier2004estimating}, MNIST handwritten digits and natural images \citep{olshausen1997sparse} and then we evaluated its performance in predicting and interpolating  3D rotation on the NORB dataset \citep{norb}.

In our test, we drew an image from one of the three databases, applied one of the transformation to generate another transformed image.  This pair of images were feed into nodes $\boldsymbol{x}_{1}$ and $\boldsymbol{x}_{2}$ of our model, which was trained using the mixture transformation sequences of natural images in the previous section. The bottom-up top-down process simultaneously infer  the latent contextual variables $\boldsymbol{z}$  and the subsequent frame  $\boldsymbol{x}_{3}$ as the prediction of the model.  In GBM, $\boldsymbol{z}$ would be first inferred, and then applied to the second frame to generate the third frame. In contrast, the prediction of the next frame in our model is not limited to the second frame, but could be based on as many previous frames as stipulated by the model.

\begin{figure}[h]
  \vspace{-0.8em}
  \centering
  \includegraphics[width=24em]{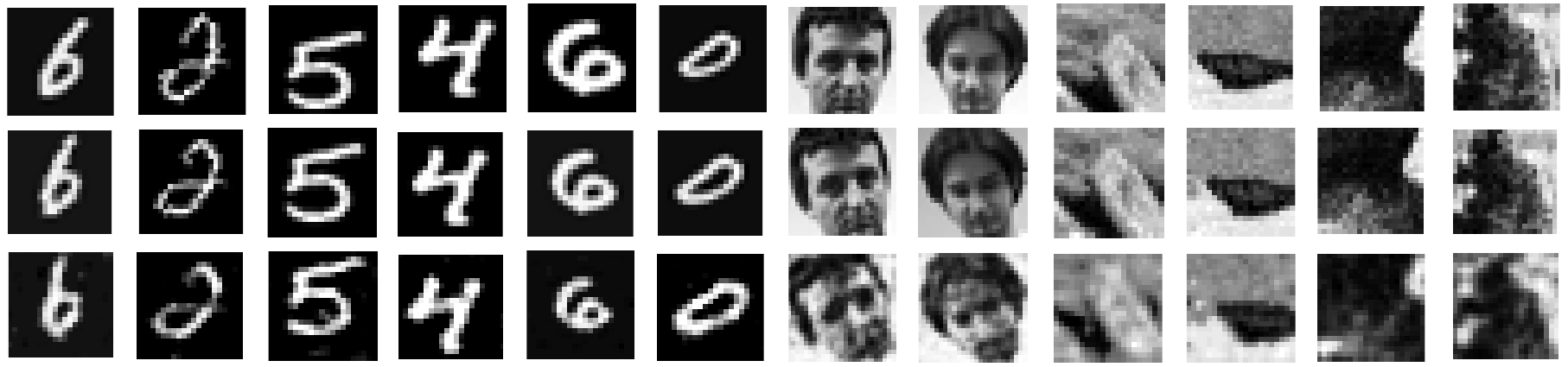}
  \vspace{-0.5em}
  \caption{Prediction results (third row) generated by the model based on the first two rows of frames.}\label{fig_pred_toy}
  \vspace{-0.5em}
\end{figure}

Figure \ref{fig_pred_toy} shows the results of predictions (third row) given the first and second frames. They demonstrate our model's ability to accomplish prediction using the  top-down bottom-up algorithm. These findings  show the contextual representation $\boldsymbol{z}$ encoded sufficient content-invariant transformation information for providing contextual modulation to generate predictions. 


We used a similar parameter setting as that in \cite{michalski2014modeling}. Each chirp sequence contained 160 frames in one second, partitioned into 16 non-overlapping 10-frame intervals, yielding 10-dimensional input vectors. The frequency of chirp signals varies between 15Hz and 20Hz. The task is to predict  the subsequent intervals given the first 5 intervals. The  RMSE results per interval for each of the subsequent intervals being predicted is shown in Table 4.

\makeatletter\def\@captype{table}\makeatother
\begin{minipage}{.4\textwidth}
\centering
\begin{tabular}{|c|c|c|}
\hline
 & Prediction & Interpolation \\
\hline
Ours & \textbf{0.2194} & \textbf{0.3510} \\
\hline
GBM & 0.2552 & - \\
\hline
GAE & 0.2281 & - \\
\hline
\end{tabular}
\caption{RMSE of prediction and interpolation (``-'' means not applicable).}\label{tab:pred_inter_accuracy}
\end{minipage}
\hfill
\makeatletter\def\@captype{table}\makeatother
\begin{minipage}{.7\textwidth}
\centering
    \begin{tabular}{|c|c|c|c|c|c|c|}
    \hline
    Predict-ahead Interval & 1     & 2     & 3     & 4     & 5     & 6 \\
    \hline
    Ours & 0.78 & 0.87  & 0.93  & 1.03  & 1.02  & 1.04  \\
    \hline
    RNN   & 0.71 & 0.85 & 0.95 & 1.04  & 1.02 & 1.03 \\
    \hline
    CRBM  & 1.00  & 1.03 & 1.01  & 1.02  & 1.04 & 1.06 \\
    \hline
    \end{tabular}%
  \caption{RMSE of prediction on chirp signals}
  \label{tab:rmse_chirp}%
\end{minipage}

\begin{figure}[h]
  \vspace{-0.5em}
  \centering
  \includegraphics[width=24em]{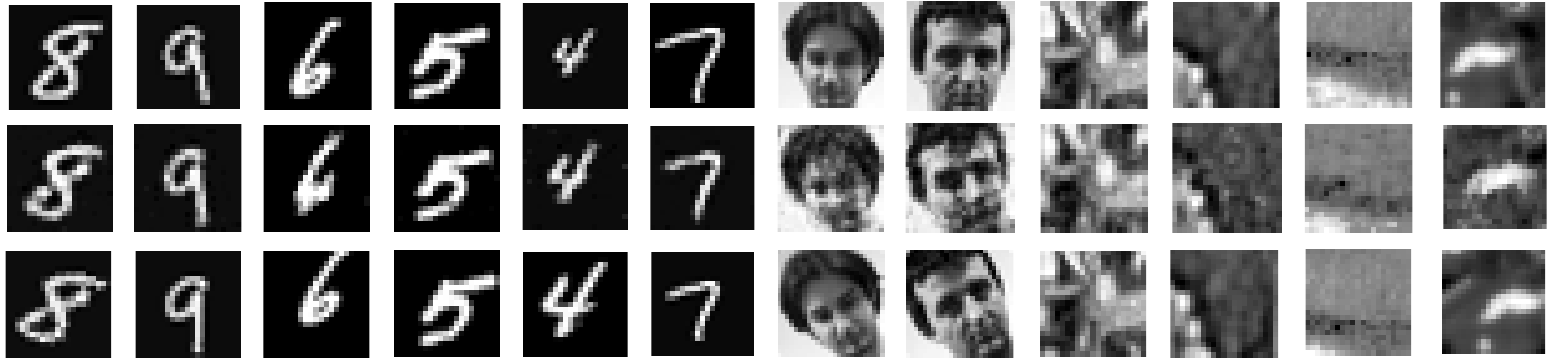}
  \vspace{-0.5em}
  \caption{Interpolation results (second row) generated based on the other two rows of frames.}\label{fig_inter_toy}
  \vspace{-1em}
\end{figure}

Interpolation can be accomplished in the same way. When  $\boldsymbol{x}_{1}$ and $\boldsymbol{x}_{3}$ are provided to the model, $z$ and $\boldsymbol{x}_{2}$ are simultaneously inferred.  The second row of Figure \ref{fig_inter_toy} shows the interpolation results; GBM cannot perform such computation.

\begin{figure}[h]
    \centering
    \vspace{-2pt}
    \begin{minipage}[b]{0.45\textwidth}
    \centering
    \fbox{\includegraphics[scale=0.8]{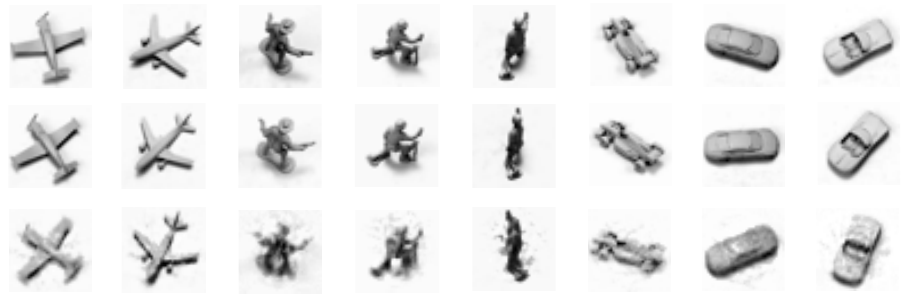}}
    \end{minipage}
    \hfill
    \begin{minipage}[t]{0.45\linewidth}
    \centering
    \fbox{\includegraphics[scale=0.55]{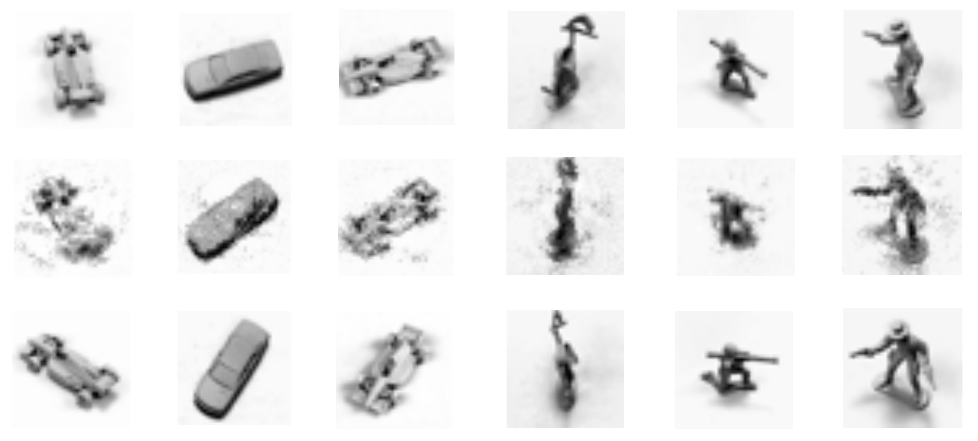}}
    \end{minipage}
    \caption{Left: Prediction results (third row) of NORB. Right: Interpolation results (second row) of NORB.}\label{fig_norb}
    \vspace{-0.7em}
\end{figure}

Next, we tested our model with a more challenging the NORB dataset that  contains  images of objects in 3D rotation, under different views and lighting conditions. There were 5 categories of objects (animals, human, cars, trucks, planes) with 10 objects in each category taken with  a camera in 18 directions, 9 camera elevations and under 6 illuminations. We trained a model for each of the five object categories. Within each category, the data were divided into a training set and a test set based on their elevations. The test set for each model included all the images taken at two particular elevations (4th and 6th), and the training set included image sequences taken at the 7 other elevations. At each elevation, the camera was fixed and the object was rotating in 3D across frames. To train the model for each category, we took sequences of three successive frames (each representing a view from a particular azimuth) of each object in under a particular condition to learn a 3-input model. We tested the model with two input images in a sequence taken from one of  the untrained elevations. The prediction results of the NORB dataset were obtained in a similar manner for the face and digit cases by presenting the image frames to $\boldsymbol{x}_{1}$ and $\boldsymbol{x}_{2}$ to infer $\boldsymbol{z}$ and $\boldsymbol{x}_{3}$ simultaneously.  The results of the prediction are shown in the third column of each object instance in  Figure \ref{fig_norb}. These results are comparable to the reported in GBM, actually with slightly better performance (see Table 3).
For all the prediction and interpolation results, we normalized the output images by matching the pixel histogram of the output image with that of the input images using histogram matching techniques.

The receptive fields of the model trained with three or more consecutive frames in this database exhibited a quadrature phase relationship between adjacent frames.  That means the filters for $\boldsymbol{y}_{1}$ and $\boldsymbol{y}_{3}$ have a phase shift of $180$ degrees.
With only the responses of these two filters to $\boldsymbol{x}_{1}$  and $\boldsymbol{x}_{3}$, but missing  $\boldsymbol{x}_{2}$, the direction of motion is underdetermined -- the movement  could go in either direction.  The model fails to interpolate in this case.  We improved the temporal resolution of the model by training the filters for $\boldsymbol{y}_{1}$ and $\boldsymbol{y}_{3}$ with the sequences, which develop a quadrature phase relationship, then we fixed the filters for $\boldsymbol{y}_{1}$ and $\boldsymbol{y}_{3}$ to train $\boldsymbol{y}_{3}$  with more sequences. This allowed the adjacent filters in the model to have finer phase difference, and yielded reasonable interpolation results, as shown in Figure \ref{fig_norb}.  A more elegant solution to this problem requires further investigation.

We  reported  the performance (root mean square error) of our model on prediction and interpolation quantitatively in Table \ref{tab:pred_inter_accuracy}. We also used 3-way GBM and GAE in the prediction test as described in Section \ref{subsec:evaluation_understand}. All the models were  trained using the mixture of motion sequences and tested on other image sequences. The result suggests that our model is comparable to, and in fact slightly outperforms, the gated machines in prediction. Additionally, our model can perform interpolation, which is not possible for GBM.

\section{Discussion}\label{sec:discussion}

In this paper, we have presented a new predictive coding framework that can learn to encode contextual relationships to modulate analysis by synthesis during perception.  As discussed in the Introduction, this model is distinct from the standard predictive coding model \citep{rao1999predictive} and in addition to being conceptually novel, might be more biologically plausible. Our model shares some similarities with the autoencoder model but differs in that (1) our model uses contextual representation to gate the prediction synthesis process while the autoencoder does not utilize context, (2) the autoencoder relies solely on fast feed-forward computation while our model utilizes a fast top-down and bottom-up procedure to update the contextual representations which modulate image synthesis during inference. Such contextual variables can be considered as a generalized form of the smoothness constraint in early vision, and can be implemented locally by interneurons in each particular visual area. A key contribution of this work is in demonstrating for the first time the usefulness of local contextual modulation in the predictive coding or the auto-encoder framework.

Recurrent neural networks still provide state-of-the-art performance in sequence modeling. But RNN requires a lot of data to train. Thus,  despite its power in modeling short and long sequences, particularly when trained with large dataset, it falters in a more limited dataset like the NORB database. The predictive encoder proposed here learns contextual latent variables that can provide information about transformation explicitly while RNN¡¯s latent variables are used to represent the content of images only and the transformations are encoded in connections.  The transforms encoded in the latent variables in the predictive encoder are directly related to the perceptual variables such as motion velocity, optical flow and binocular disparity.

Our model shares similar goals with the gated Boltzmann machine in learning transformational relationships but our model utilizes a different mechanism, with a unified framework for inference, interpolation and prediction. We consider the GBM as a state-of-art method for learning transform with limited amount of data and thus we have mostly focused our quantitative evaluation of our predictive encoder model against the performance of GBM. We found that our model is comparable or superior in performance relative to the gated Boltzmann machine in terms of inference and prediction while being additionally able to perform interpolation. Our model relies on standard  spatiotemporal filtering in the feedforward path, without the need for the N-way multiplicative interaction or neural synchrony as required in the GBM. It is thus simpler in conceptualization and maybe more biologically plausible. It is important to recognize that our model currently just is a module that uses latent variables to encode spatiotemporal local context and transformation. Such a module can be used to build recurrent neural networks to model the temporal evolution of the contextual variables, or be stacked up  to form deep networks to learn hierarchical features, each layer with their own local spatiotemporal contextual representations.

\section{Acknowledgments}\label{sec:ack}
\vspace{-1em}
This research  was supported by research grants 973-2015CB351800, NSFC-61272027 (to YZ Wang), and NSF 1320651 and NIH R01 EY022247 (to TS Lee). Both Wang and Lee's labs acknowledge the support of NVIDIA Corporation for the donation of GPUs  for this research. Mingmin Zhao and Chengxu Zhuang were supported by Peking University and Tsinghua University undergraduate scholarships  respectively when they visited Carnegie Mellon to carry out this research.

\bibliographystyle{iclr2015}
\bibliography{bib/bib_file}
\end{document}